\documentclass{article}

\usepackage{PRIMEarxiv}

\usepackage[utf8]{inputenc} 
\usepackage[T1]{fontenc}    
\usepackage{hyperref}       
\usepackage{url}            
\usepackage{booktabs}       
\usepackage{amsfonts}       
\usepackage{nicefrac}       
\usepackage{microtype}      
\usepackage{lipsum}
\usepackage{fancyhdr}       
\usepackage{graphicx}       
\graphicspath{{media/}}     
\usepackage{subcaption}

\usepackage{xcolor}

\title{To be or not to be? An exploration of continuously controllable prompt engineering
\thanks{} 
}

\author{
  Yuhan Sun \\
  Zhejiang University \\
  SenseTime \\
  Hangzhou, China \\
  \texttt{sunyuhan@zju.edu.cn} \\
   \And
  Mukai Li \\
  Hongkong Univeristy \\
  Hongkong, China \\
  \texttt{limukaikiaia@163.com} \\
   \And
  Yixin Cao \\
  Singapore Management University \\
  Singapore \\
  \texttt{yxcao@smu.edu.sg} \\
   \And
  *Kun Wang \\
  SenseTime \\
  Beijing, China \\
  \texttt{wangkun@sensetime.com} \\
   \And
  Wenxiao Wang \\
  Zhejiang University \\
  Hangzhou, China \\
  \texttt{wenxiaowang@zju.edu.cn} \\
   \And
  Xingyu Zeng \\
  SenseTime \\
  Shenzhen, China \\
  \texttt{zengxingyu@sensetime.com} \\
   \And
  Rui Zhao \\
  SenseTime \\
  Shenzhen, China \\
  \texttt{zhaorui@sensetime.com} \\
}

\begin{document}

\newcommand{\mukai}[1]{\textcolor{orange}{\bf \small [#1 --mukai]}}

\maketitle

\begin{abstract}
As the use of large language models becomes more widespread, techniques like parameter-efficient fine-tuning and other methods for controlled generation are gaining traction for customizing models and managing their outputs. However, the challenge of precisely controlling how prompts influence these models is an area ripe for further investigation. In response, we introduce ControlPE (Continuously Controllable Prompt Engineering). ControlPE enables finer adjustments to prompt effects, complementing existing prompt engineering, and effectively controls continuous targets. This approach harnesses the power of LoRA (Low-Rank Adaptation) to create an effect akin to prompt weighting, enabling fine-tuned adjustments to the impact of prompts. Our methodology involves generating specialized datasets for prompt distillation, incorporating these prompts into the LoRA model, and carefully adjusting LoRA’s merging weight to regulate the influence of prompts. This provides a dynamic and adaptable tool for prompt control. Through our experiments, we have validated the practicality and efficacy of ControlPE. It proves to be a promising solution for control a variety of prompts, ranging from generating short responses prompts,refusal prompts to chain-of-thought prompts.
\footnote{Work in progress}
\end{abstract}


\section{Introduction}
\label{sec:intro}

Recently, large language models (LLMs) have achieved great success in both industrial and academia applications, such as Med-PaLM\cite{singhal2022large}, ChatLaw\cite{cui2023chatlaw} and LLMs for education\cite{KASNECI2023102274}. 
They not only unify various tasks in the form of text generation, but also eliminate the need of task-specific training via a new paradigm "pre-train, prompt then predict". A key challenge lies in how to design suitable prompts. Evidence shows that a subtle modification of prompt may lead to significant performance change\cite{liu2023gpt}.  
Among almost unlimited number of natural language prompts, it is difficult to identify the optimal ones. Therefore, prompt engineering has received much attention to ease the discovery. 

Although many existing prompt optimization techniques have been proposed to automate the prompt engineering procedure, few work focuses on further controlling or combining these discrete prompts in a continuous space. APE(Automatic Prompt Engineer)\cite{zhou2023ape}, a novel approach for automatically generating and selecting effective prompts to guide LLMs). It treats the creation of prompts as a form of program synthesis, where an LLM generates a pool of potential prompts. PromptAgent\cite{wang2023promptagent} treats prompt optimization as a strategic planning problem. This means it systematically navigates through the space of potential prompts, aiming to find the most effective ones. However, these automatic prompt engineering methods struggle to find an intermediary state between adding and not adding the already retrieved prompts, making it challenging to identify a balanced approach.

In this paper, we propose a novel continuously controllable prompt engineering method, ControlPE. It can not only further adjust the effects of a prompt at a finer grained level, complementary to existing prompt engineering works, but also is particularly helpful in controlling continuous targets. The previous automated prompt engineering tasks still revolve around searching for the most suitable prompt within discrete natural language prompts. The use of discrete prompts in prompt engineering presents significant limitations. For instance, incorporating a prompt like "keep the answer short and concise" can lead to a 55\% reduction in the model's average response length. Such a reduction, though significant, can result in the loss of information in responses. Therefore, if we aim to further shorten the model's average response length to 20\% or 30\%, achieving this solely through discrete prompt searching proves to be quite challenging.

To achieve ControlPE, we present a three-step methodology: first, we generate targeted prompt distillation\cite{sahu2023promptmix,anil2020large} datasets, then train LoRA\cite{hu2021lora} models on these datasets in purpose of distilling the prompt into the LoRA parameters, and finally, we tune LoRA's merging weight\cite{yang2023adamerging,zhang2023composing} to achieve prompt weighting. Our experiments reveal the utility of this approach in optimizing model responses, be it for providing short and concise answers or handling refusal prompts. This research offers a flexible and powerful tool for tailoring model behavior, paving the way for more nuanced and context-aware responses in natural language processing.

Our contribution can be summarized as follows:

\begin{itemize}
    \item  We identify and highlight the needs of continous variable in prompt optimization, complementary to existing prompt engineering techniques. 
    \item As we know first work to propose a methodology that enables continuous prompt engineering. This method allows for finer control over prompt effects, introducing a new dimension of flexibility in LLMs.
    \item Implement diverse scenarios such as controlling the response length, providing refusal answers in Document-Based Question Answering (DocQA) tasks, and utilizing chain-of-thought reasoning in mathematical problem-solving tasks.
\end{itemize}

\begin{figure}[ht!]
    \centering
    \includegraphics[width=0.8\textwidth]{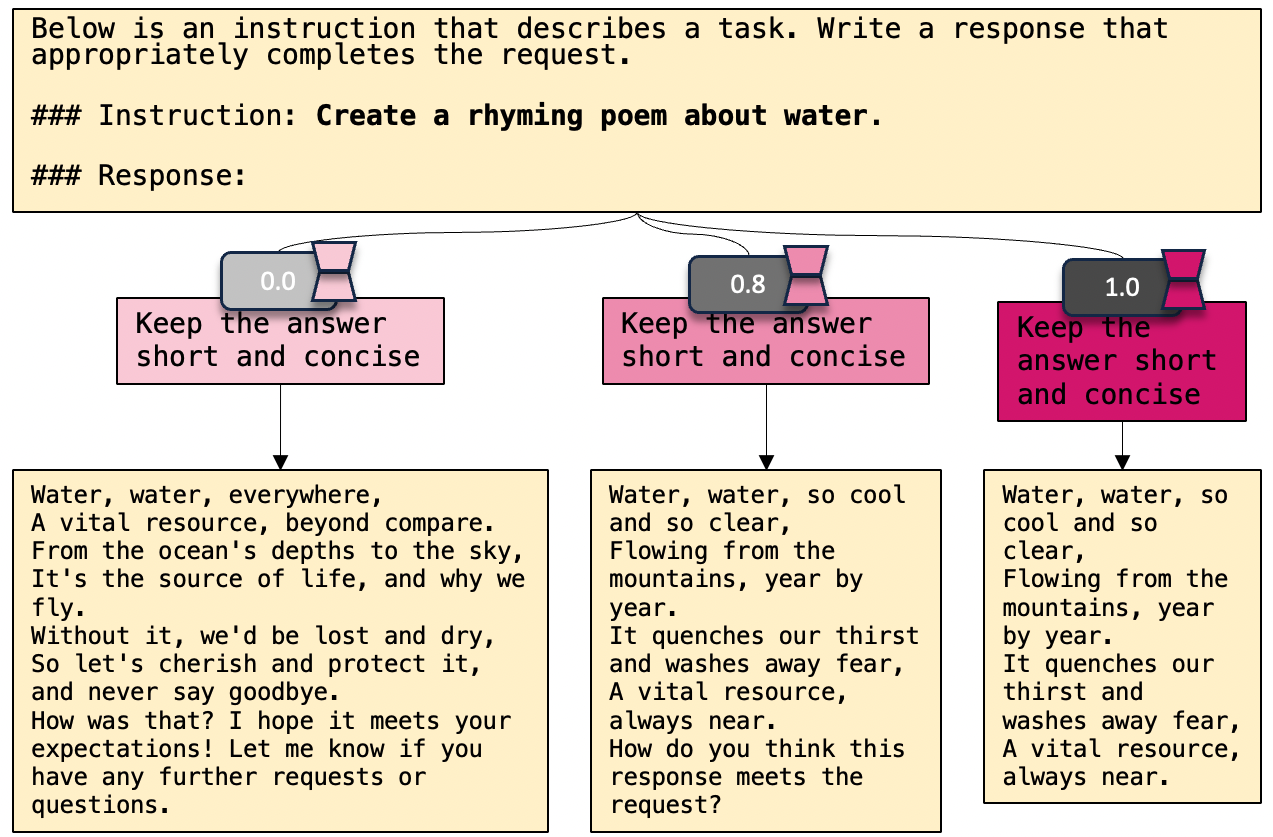}
    \caption{Effect demonstration picture of the ControlPE applied to "keep the answer short and concise"}
    \label{fig:example}
\end{figure}
\begin{figure}[ht!]
    \centering
    \includegraphics[width=0.8\textwidth]{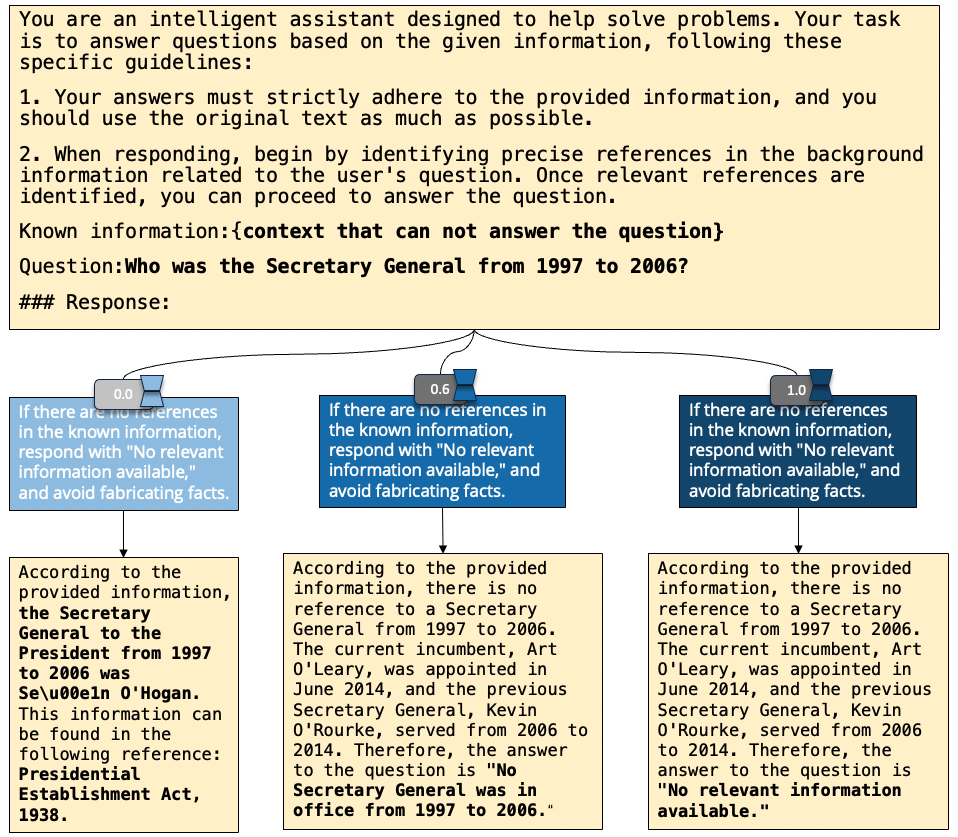}
    \caption{Effect demonstration picture of the ControlPE applied to "If there are no references in the known information, respond with "No relevant information available," and avoid fabricating facts"}
    \label{fig:example}
\end{figure}

\section{Related works}

\subsection{Prompt Engineering}
Prompt engineering is a key method for interacting with generalist models like LLMs, offering a user-friendly way for people to use these technologies. This approach has become popular for various NLP tasks, as highlighted in studies by \cite{schick2021exploiting,brown2020language,sanh2022multitask}. However, designing effective prompts for LLMs is a nuanced task. It can be done either manually, as seen in \cite{reynolds2021prompt}, or through automated systems, as in the works of \cite{gao2021making,shin2020autoprompt}. This is because LLMs don't interpret prompts as humans do, a challenge noted by \cite{webson2022promptbased,lu2022fantastically}.

Many successful approaches to prompt tuning involve gradient-based optimization in a continuous space, as discussed in studies by \cite{liu2023gpt,qin2021learning,lester2021power}. However, this method becomes less feasible at larger scales due to the high computational cost and the shift towards using API-accessible models that don't provide gradient information.

For instance, a prompt such as "keep the answer short and concise" can effectively reduce the length of a model's responses by an average of 55\%. This becomes particularly significant in document-based question-answering (QA) tasks, where it is crucial to manage the model's tendency to produce information not present in the source material. By incorporating prompts like "If there are no references in the known information, respond with 'No relevant information available,' and avoid fabricating facts," the model can be guided to refuse to answer when the context does not contain information pertinent to the question. Similarly, prompts designed to encourage step-by-step reasoning\cite{wei2023chainofthought}, such as "Let's think step by step," have shown to enhance the model's performance on mathematical datasets. The effectiveness of prompt engineering underscores its value in refining the utility and accuracy of language model interactions.
\subsection{Controllable Text Generation}
Conventional decoding algorithms are limited in that they cannot integrate constraints during the generation process, making it challenging to ensure that the output meets specific requirements crucial for some applications. In response, research has branched into two main areas focusing on controlled text generation: constrained search algorithms and score-based sampling methods.

Constrained search algorithms work by imposing strict lexical constraints on the outputs, altering the search space to adhere to these constraints. An example is the Constrained Beam Search (CBS) algorithm by \cite{anderson2017guided}, which uses a finite-state automaton to track constraint satisfaction. However, CBS has a significant drawback in its complexity, as it requires maintaining a finite-state machine with $2^C$ states (C being the number of constraints), leading to increased time complexity. To address this, \cite{hokamp-liu-2017-lexically} and \cite{post-vilar-2018-fast} introduced Grid Beam Search (GBS) and Dynamic Beam Allocation (DBA), respectively. Furthermore, \cite{lu-etal-2021-neurologic} developed NeuroLogic decoding, a constrained search algorithm for satisfying complex logic-based lexical constraints in conjunctive normal form, later enhanced by NeuroLogic A* decoding \cite{lu-etal-2022-neurologic} with lookahead heuristics. While these algorithms improve constraint satisfaction rates, they often result in slower generation speeds and lower quality text due to aggressive pruning of the output distribution space, leading to suboptimal results. Additionally, most of these algorithms are limited to lexical constraints.

On the other hand, score-based sampling methods incorporate constraints by transforming them into differentiable score functions. Soft constraints, like sentiment control, can be implemented using the cross-entropy loss of classifiers, while hard constraints, such as lexical constraints, can be modeled by a differentiable n-gram matching function \cite{liu-etal-2022-dont}. These methods are more versatile than constrained search algorithms, handling a wider variety of constraints and their combinations. However, they lack guarantees for constraint satisfaction and often compromise generation quality due to alterations in the output distribution \cite{qin2022cold}. Additionally, these methods are slower due to the need for multiple score-matching steps and require careful tuning of the weights between different constraints and task-specific losses to balance output quality and constraint satisfaction.

Beyond these two methods, \cite{dinu-etal-2019-training} proposed a specialized training approach that includes lexical constraints in the input during model training. Similarly, \cite{keskar2019ctrl} is pre-trained with structures that naturally co-occur with raw texts, allowing it to handle constraints related to style and domain.

\section{Method}
\label{sec:headings}

\begin{figure}[ht!]
    \centering
    \includegraphics[width=1.0\textwidth]{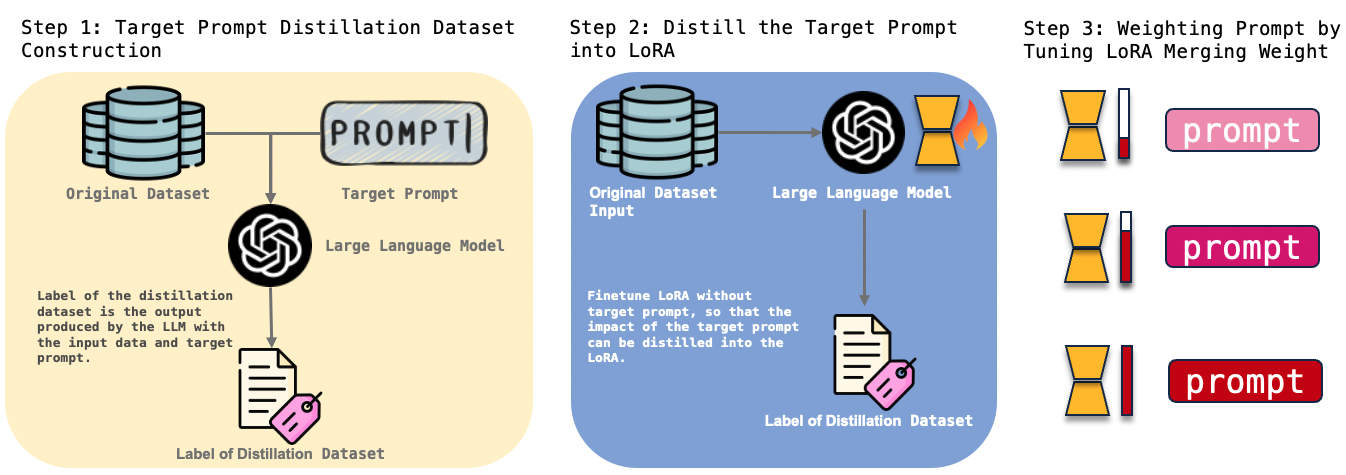}
    \caption{Three steps to achieve prompt weighting via LoRA}
    \label{fig:example}
\end{figure}
This section outlines the method used to achieve prompt weighting via LoRA in three steps. As illustrated in Figure 3, the first step involves utilizing instruction data augmented with a target prompt for model inference, which yields labels for constructing a distillation dataset. In the second step, we train the LoRA model using instruction data without adding the target prompt, thereby distilling the influence of the prompt into the LoRA model. In the third step, we fine-tune the influence of the prompt by adjusting the weights of the LoRA model.
\subsection{Preliminary}
\paragraph{LoRA}
LoRA\cite{hu2021lora} stands out as a parameter-efficient fine-tuning\cite{peft} methods large language models, particularly in scenarios with limited computational resources. This approach has gained traction, as evidenced by its inclusion in state-of-the-art models like LLaMA\cite{touvron2023llama,touvron2023llama2}, and it's highlighted for its efficiency in recent literature. LoRA functions by adjusting the output of transformer weight matrices. Typically, these matrices convert an input $x$ from one dimension to another $h$, but with LoRA, this conversion is tweaked by adding the product of two projection matrices, $B$ and $A$, with $x$. These matrices are much smaller, which reduces complexity since $r$, the rank, is much smaller than the dimensions $d$ and $k$. This modification is primarily applied to the query and value projection matrices within the attention mechanism of transformers. When initiating LoRA tuning, the matrix $A$ is filled with random Gaussian values, while $B$ starts with zeros. This setup ensures that the original pre-trained model's behavior is preserved at the start. The combination of $A$ and $B$ comprises the LoRA module, which can be integrated with other distinctively trained LoRA modules for enhanced performance.
\subsection{Step 1: Target Prompt Distillation Dataset Construction}

In the initial phase, we will embark on training a LoRA that is functionally equivalent to the target prompt. In essence, our objective is to distill the prompt's influence on the model into LoRA. Aiming to achieve this effect, we begin by constructing a target prompt distillation dataset. This dataset serves as the foundation for training LoRA models to represent specific prompts. 
\begin{equation}
y_{target} = \theta(P_{original}, P_{target}, x)
\end{equation}
\begin{equation}
D_{distill}=\{(x,y_{target})\}_n
\end{equation}
The process involves using input prompts that incorporate the target prompt, performing inference, and constructing a new dataset $D_{distill}$ with input data $x$ inherited from the original dataset and labels $y_{target}$ are generated by the large language model $\theta$ using the original prompt template $P_{original}$, target prompt $P_{target}$ injected with input data $x$ as input.

\subsection{Step 2: Distill the Target Prompt into LoRA}
The next step involves training LoRA models to distill the target prompts. During this training phase, we remove the target prompts from the input prompts.

\begin{equation}
\theta_{LoRA} = Apply(\theta,LoRA)
\end{equation}
\begin{equation}
L_{LoRA} = \sum_{(x,y_{target})\in D_{distill}} CrossEntropy(\theta_{LoRA}(P_{original}, x), y_{target})
\end{equation}
We demonstrate the process of applying LoRA to the model's parameters $\theta$. This results in the creation of a modified model, denoted as $\theta_{LoRA}$, which incorporates the influence of LoRA. \\
$L_{LoRA}$ denoted as the loss function of LoRA, that assesses the disparity between the model's predictions and the target labels generated in the last step. It operates on a dataset $D_{distill}$, where each data point consists of input $x$ and its corresponding target label $y_{target}$. The $CrossEntropy$ function quantifies the dissimilarity between the model's prediction and the target label for each data point. The objective is to minimize this loss, guiding the model to make LoRA approximate the impact of the target prompts. In essence, LoRA is fitted to the target prompts in this step.

\subsection{Step 3: Weighting Prompt by Tuning LoRA Merging Weight}
In the final step, we introduce prompt weighting by tuning LoRA's merging\cite{yang2023adamerging,zhang2023composing} weight. Having already obtained LoRA models that approximate the target prompts, we fine-tune LoRA's parameter weights to control its actual influence on the model. As a result, the prompt's impact can be adjusted by dynamically modifying the LoRA parameter weights. 
\begin{equation}
\theta_{\alpha LoRA} = Apply(\theta,\alpha LoRA)
\end{equation}
\begin{equation}
\theta(P_{original},\alpha P_{target}) \approx \theta_{\alpha LoRA}(P_{original})
\end{equation}
This approach allows for precise control over prompt influence, and its effectiveness can be empirically verified by demonstrating the validity of equations (6) as presented in the method.
\section{Experiments}
\label{sec:others}
In this section, we present the experiments conducted to evaluate the effectiveness of our proposed method for weighting prompts using LoRA in the field of Natural Language Processing.


\subsection{Experimental Settings}
Our foundational model is LLaMA2-7B-chat\cite{touvron2023llama2}, and we utilize the peft\cite{peft} library for training all LoRA adapters. These adapters are integrated into every q\_proj and v\_proj module within the transformer architecture. We've configured the LoRA rank at 16. For our training process, we employ a batch size of 128 and a learning rate of 3e-4. The entire distillation training is conducted over a span of 3 epochs.
\subsection{ControlPE on Response Length Control}

\begin{figure}
    \centering
    \begin{subfigure}{0.45\textwidth}
        \includegraphics[width=1\textwidth]{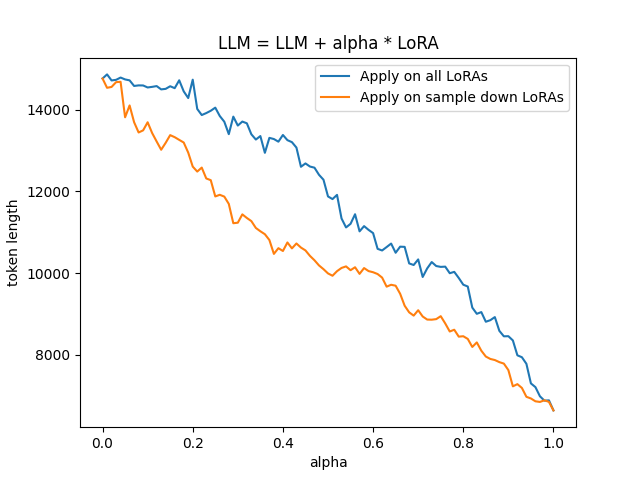}
        \caption{The response length of a language model varies with adjustments to the LoRA weights}
        \label{fig:image1}
    \end{subfigure}
    \begin{subfigure}{0.45\textwidth}
        \includegraphics[width=1\textwidth]{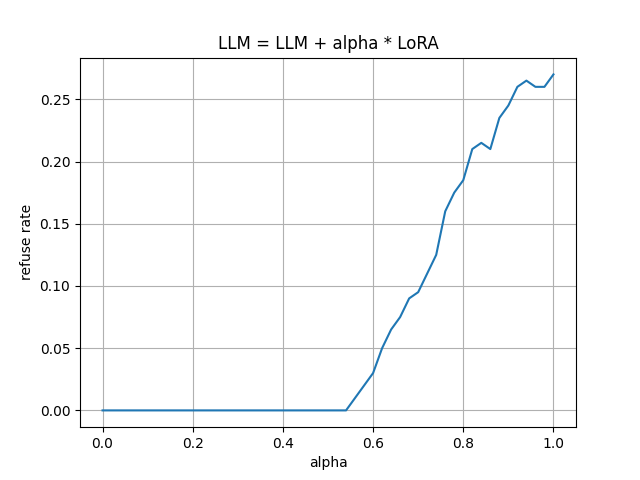}
        \caption{The refusal-to-answer rate of language models changes with adjustments to LoRA weights}
        \label{fig:image2}
    \end{subfigure}
    \caption{Figure (a) illustrates the variation in response length of large language models with adjustments in LoRA weights, while Figure (b) depicts the change in refusal-to-answer rate of language models with LoRA weight adjustments. In Figure (a), the yellow line represents applying weights only to the 'sample down' LoRA matrix, whereas the blue line represents applying weights to all matrices in LoRA. In Figure (b), weights are exclusively applied to the 'sample down' LoRA.}
    \label{fig:example}
\end{figure}

Firstly, we demonstrated the effectiveness of ControlPE in controlling the response length of a model. 
Our test dataset comprised instructions numbered 10001 to 10100 from the 52k alpaca-data\cite{alpaca}. When employing the standard alpaca response prompt template, the total response length for these 100 instructions was 14,766 tokens. However, after integrating the prompt "Keep the answer short and concise" into the standard template, the total response length for the same set of instructions was reduced to 7,096 tokens. We then used this modified prompt template with the added "Keep the answer short and concise" on the first 10,000 instructions from the 52k alpaca-data for inference with the original model. The inference results were used as labels for a distillation dataset, and the standard prompt template without the "Keep the answer short and concise" was used as input for distillation training. Through this process, we aimed to distill the influence of the target prompt into LoRA parameters. The LoRA model, post-distillation, produced a total response length of 6,640 tokens on the 100 test instructions.

Subsequently, as shown in Figure 4(a) we adjusted the weights in the LoRA model and observed its performance on the 100 test instructions. We found that adjusting the weights of the LoRA matrix can effectively control the model's response length. Applying weights only to the sample down LoRA matrix allows for a linear adjustment of the model’s response length. Applying weights to both matrices results in a non-linear relationship between the response length and the weights. Therefore, we recommend applying weights to only a single LoRA matrix for all experiments.

\subsection{ControlPE on Refusal Answer}

\begin{figure}[ht!]
    \centering
    \includegraphics[width=0.6\textwidth]{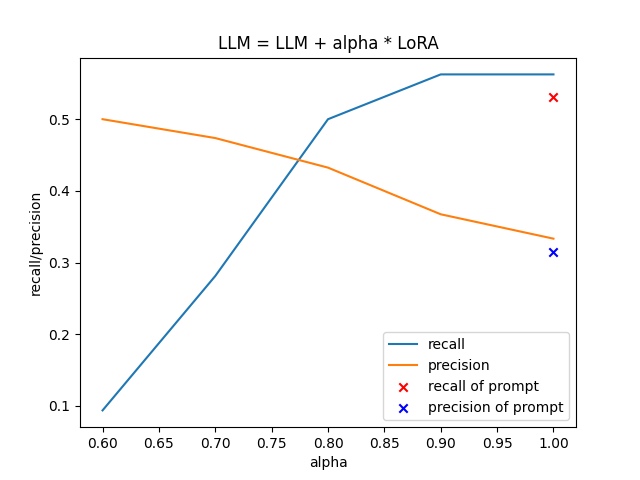}
    \caption{Effect demonstration picture of the ControlPE applied to "If there are no references in the known information, respond with "No relevant information available," and avoid fabricating facts"}
    \label{fig:example}
\end{figure}

In this section, we will demonstrate how the ControlPE performs in refusing to answer questions in the DocQA\cite{ho2021foldem} task. The DocQA task involves a language model answering questions based on a provided document. However, in real-world scenarios, there are instances where the document cannot answer the question, necessitating a prompt that guides the model to refuse to answer - "If there are no references in the known information, respond with 'No relevant information available,' and avoid fabricating facts." We utilized the wikipedia-trivia\cite{wikipedia-trivia} dataset, specifically entries 10001 to 10200, where 80\% of the documents were positive (meaning the given context was sufficient to answer the question, and a direct answer was expected rather than a refusal) and 20\% were negative (meaning the context was insufficient, and the model should respond with "No relevant information available"). Before adding the refusal prompt, the refusal rate was 0\%. After adding the refusal prompt, the refusal rate increased to 27\%, as indicated by the red crosses in the graph. Within this 27\% refusal rate, the precision (the accuracy of refused answers among questions that should be refused) was 31\%, and the recall (the rate of questions that should be refused and were indeed refused) was 53\%.

We used the first 10,000 entries of the wikipedia-trivia dataset with the refusal prompt added to train the LoRA model, treating the inference results as labels. This method distilled the refusal capability of the prompt into LoRA. The refusal rate of the distilled model on the test set was 29\%. As shown in the graph, by adjusting the weight, the refusal rate of the model linearly increased with weights between 0.6 to 1.0. In the 0.1-0.5 weight range, the model did not directly refuse to answer but reduced hallucinations and indirectly refused many questions (we consider a direct refusal only when the model explicitly states "No relevant information available").

Furthermore, by adjusting the weight between 0.6 and 1.0, we tested the model's recall and precision in answering. We observed that as the refusal capability strengthened, precision weakened while recall increased. Therefore, we can find an appropriate trade-off between recall and precision through parameter tuning.

\subsection{ControlPE on CoT}

\begin{figure}[ht!]
    \centering
    \includegraphics[width=0.6\textwidth]{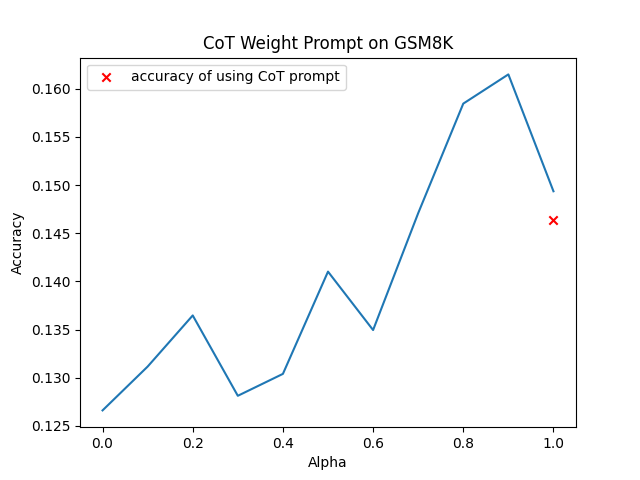}
    \caption{Effect demonstration picture of the ControlPE applied to "Let's think step by step"}
    \label{fig:example}
\end{figure}

In this experiment, we directly utilized OpenAI's GSM8K\cite{cobbe2021training} dataset for ControlPE in Chain of Thought (CoT)\cite{wei2023chainofthought} analysis. We observed that without the prompt "Let's think step by step," the accuracy of Llama2-7b-chat on the GSM8K dataset was 12.6\%. However, adding the prompt "Let's think step by step" increased the model's accuracy to 14.6\%. We then constructed a CoT distillation dataset on the first 10k instructions from the 52k alpaca dataset: we used inferences with the added prompt "Let's think step by step" as labels and instructions without this prompt as inputs. The "Let's think step by step" was distilled into the LoRA parameters. The distilled model exhibited a performance of 14.9\% on GSM8K.

Subsequently, we adjusted the weights of LoRA and observed its performance on the GSM8K test set. Surprisingly, the best performance on the GSM8K dataset was not achieved by fully applying the CoT influence on the model. Instead, according to figure 6, a slightly weaker influence, such as 80\%, led to the optimal performance of Llama2-7B-chat on GSM8K.


\subsection{Multiple Prompts Fusion Engineering}
\begin{figure}[ht!]
    \centering
    \begin{subfigure}{0.8\textwidth}
        \includegraphics[width=1\textwidth]{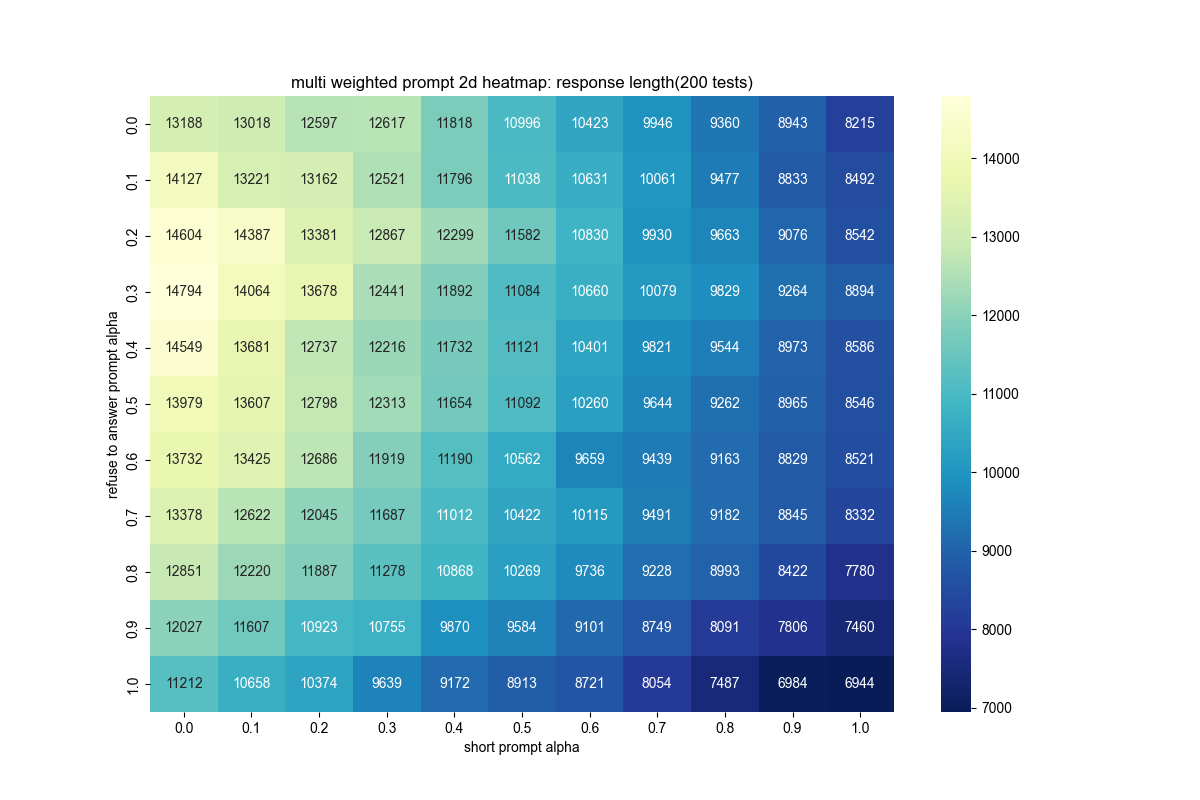}
        \caption{The response length of a language model varies with adjustments to the 2 LoRAs weights}
        \label{fig:image1}
    \end{subfigure}
    \begin{subfigure}{0.8\textwidth}
        \includegraphics[width=1\textwidth]{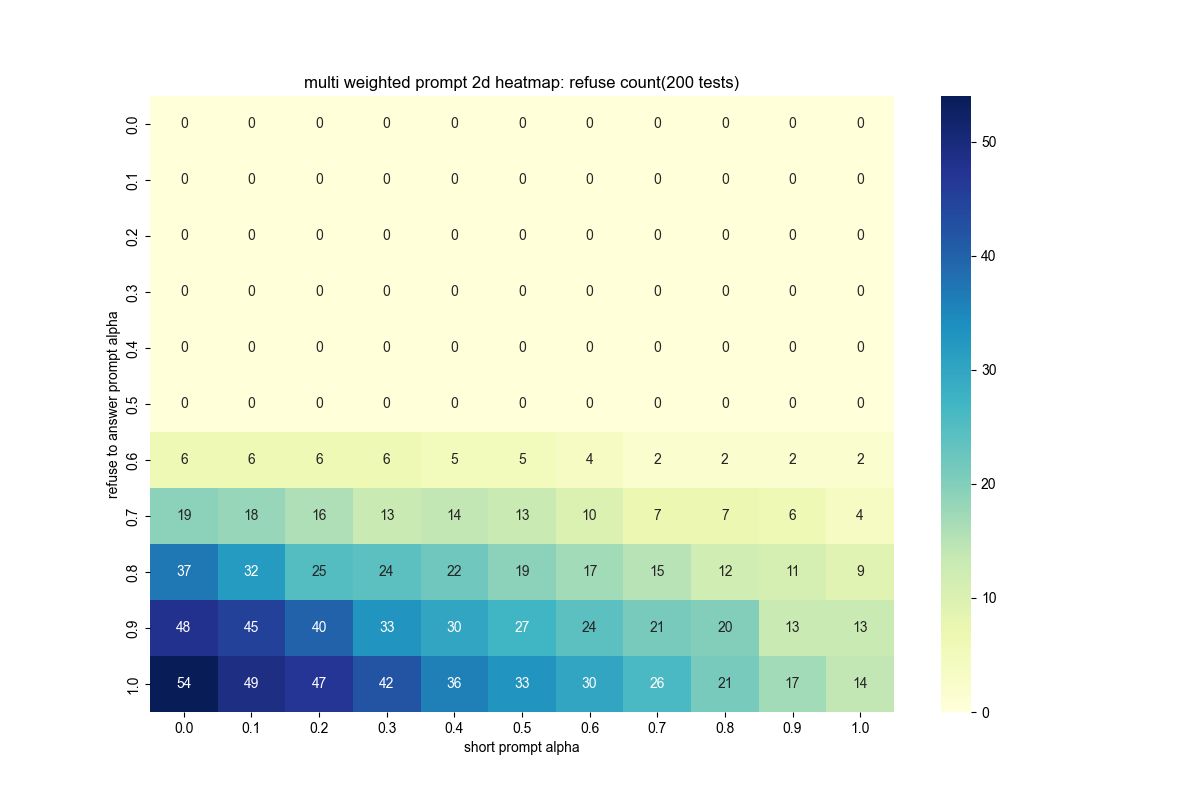}
        \caption{The refusal-to-answer count of language models varies with adjustments to the 2 LoRAs weights}
        \label{fig:image2}
    \end{subfigure}
    \caption{Figure (a) illustrates the variation in response length of large language models with adjustments in 2 LoRAs weights, while Figure (b) depicts the change in refusal-to-answer rate of language models with 2 LoRAs weight adjustments. All weights are applied to the 'sample down' LoRA.}
    \label{fig:example}
\end{figure}
In the fusion experiment, we will combine the LoRA model previously trained on alpaca-data, which was distilled with the directive "Keep the answer short and concise," with the LoRA model distilled on wikipedia-trivia with the guideline "If there are no references in the known information, respond with 'No relevant information available,' and avoid fabricating facts." Note that in this experiment, we will not train a new model but will directly use the models distilled from the two previous experiments. The test dataset used is the wikipedia-trivia DocQA data from entries 10001 to 10200, with 80\% of the documents being positive and 20\% negative.

As illustrated in the Figure 7(a) and Figure 7(b), although the LoRA model distilled with "Keep the answer short and concise" was not trained on wikipedia-trivia, it still maintains the ability to linearly regulate the response length. During the fusion, while they influence each other, they still retain the capability to adjust their influence on the prompt itself.





\section{Conclusion}
Our research presents an innovative approach to prompt weighting in natural language processing through the application of Low-Rank Adaptation (LoRA). This study is motivated by the growing importance of prompt engineering in controlling large language models, where precise control of prompt influence is crucial. By leveraging the efficiency of LoRA, our method introduces a sophisticated, yet accessible means of tuning the influence of specific prompts on model behavior.

We explored three key areas in our experiments: response length control, refusal to answer, and chain of thought reasoning. The results show that our approach can effectively modulate the influence of prompts in each scenario. For instance, in controlling response length, our method demonstrated the ability to adjust model outputs linearly, offering a fine-grained control over verbosity. In refusal-to-answer scenarios, we successfully increased the refusal rate while balancing precision and recall, addressing the challenge of information fabrication. Furthermore, in chain of thought reasoning, our method improved accuracy by selectively applying prompt influence, underscoring the importance of nuanced prompt control.

Additionally, our experiments on multiple ControlPEs fusion showcased the potential of our method in managing multiple prompts simultaneously, maintaining the ability to independently adjust each prompt's influence.

In conclusion, this research contributes significantly to the field of prompt engineering in NLP. The ability to weight prompts using LoRA not only enhances the control over large language models but also opens up new avenues for research and practical applications. Our method stands as a testament to the evolving landscape of NLP, where user control and model flexibility are increasingly paramount.

\section*{Acknowledgments}
TODO

\bibliographystyle{unsrt}  
\bibliography{references}

\end{document}